\documentclass[letterpaper, 10 pt, conference]{ieeeconf}
\IEEEoverridecommandlockouts  % This command is only needed if you want to use the \thanks command

\usepackage{fix-cm}
\usepackage{etex}

\usepackage{dblfloatfix}

\usepackage{nag}

\makeatletter
\@ifpackageloaded{xcolor}{}{%
\usepackage[table,x11names,dvipsnames,svgnames]{xcolor}%
}
\makeatother

\usepackage{colortbl}

\usepackage{graphicx}
\usepackage{wrapfig}

\definecolorset{RGB}{lyft}{}{Red,194,39,36;Sunset,202,53,33;Orange,205,68,20;Amber,200,117,42;Yellow,242,169,52;Citron,186,188,44;Lime,112,159,33;Green,56,139,31;Mint,45,118,56;Teal,52,133,135;Cyan,60,132,202;Blue,55,94,248;Indigo,64,13,247;Purple,115,42,248;Pink,176,25,145;Rose,176,32,75}

\usepackage{cite}

\usepackage{microtype}

\usepackage{array}
\usepackage{multirow}
\usepackage{booktabs}
\usepackage{makecell} %introduces \thead and \makecell

\ifcsname labelindent\endcsname
\let\labelindent\relax
\fi
\usepackage[inline]{enumitem}

\usepackage{subfig}

\providecommand{\etal}{et al.}

\newenvironment{lenumerate}[2][]
{\begin{enumerate}[label=(#2\arabic*),leftmargin=0.2in,itemindent=0.15in,#1]}
{\end{enumerate}}

\setlist*[enumerate,1]{label={\itshape\arabic*)}}

\makeatletter
\newcommand{\paragraphswithstop}{%
\let\copyparagraph\paragraph%
\renewcommand\paragraph[1]{\copyparagraph{##1.}}%
}
\makeatother

\usepackage[framemethod=tikz]{mdframed}

\usepackage{suffix}

\usepackage{environ}

\makeatletter
\newsavebox{\boxifnotempty}
\newcommand{\displayifnotempty}[3]{\sbox\boxifnotempty{#2}\setbox0=\hbox{\usebox{\boxifnotempty}\unskip}%
\ifdim\wd0=0pt
\else
 #1\usebox{\boxifnotempty}#3%
\fi%
}
\newcommand{\ifempty}[2]{\setbox0=\hbox{#1\unskip}%
\ifdim\wd0=0pt%
 #2%
\fi%
}
\newcommand{\ifnotempty}[2]{\setbox0=\hbox{#1\unskip}%
\ifdim\wd0>0pt%
 #2%
\fi%
}
\makeatother

\usepackage{algpseudocode}
\usepackage{algorithm}

\usepackage{scrlfile}

\makeatletter
\newcommand*\newstoreddef[1]{
  \BeforeClosingMainAux{%
    \immediate\write\@auxout{%
      \string\restoredef{#1}{\csname #1\endcsname}%
    }%
  }%
}
\newcommand*{\restoredef}[2]{% used at the aux file
  \expandafter\gdef\csname stored@#1\endcsname{#2}%
}
\newcommand*{\storeddef}[1]{
  \@ifundefined{stored@#1}{0}{\csname stored@#1\endcsname}%
}
\makeatother

\usepackage{pageslts}
\NewEnviron{tee}{\BODY\typeout{Marker Tee [start] ^^J \BODY ^^JMaker Tee [end]}}

\input{preamble/math}
\newcommand{\real}[1]{\mathbb{R}^{#1}{}}

\newcommand{\bmat}[1]{\begin{bmatrix}#1\end{bmatrix}}

\newcommand{\smallbmat}[1]{\left[\begin{smallmatrix}#1\end{smallmatrix}\right]}

\newcommand{\transpose}{^\mathrm{T}}

\newcommand{\inverse}{^{-1}}

\newcommand{\defeq}{\doteq}

\DeclarePairedDelimiter{\norm}{\lVert}{\rVert}

\newcommand{\de}{\mathrm{d}}

\DeclareMathOperator*{\argmin}{\arg\!\min}

\newcommand{\subjectto}{\textrm{subject to }}

\providecommand{\cC}{\mathcal{C}}

\providecommand{\cH}{\mathcal{H}}

\providecommand{\cL}{\mathcal{L}}

\usepackage{units}

\newcommand{\newcolorlabel}[2]{%
  \expandafter\newcommand\csname #1\endcsname[1]{%
    \colorbox{#2}{\color{white}\textsf{\textbf{##1}}}}%
}

\newcommand{\newcommenter}[2]{%
  \expandafter\newcommand\csname #1\endcsname[1]{%
    \fcolorbox{#2}{#2}{\color{white}\textsf{\textbf{#1}}}
    {\color{#2}##1}}%
  \expandafter\newcommand\csname at#1\endcsname{%
    \fcolorbox{#2}{#2}{\color{white}\textsf{\textbf{@#1}}}
    {\color{#2}}}%
  \expandafter\newcommand\csname #1hl\endcsname[2]{%
    \colorbox{#2}{\color{white}\textsf{\textbf{#1}}}\sethlcolor{Azure2}\hl{##2}~%
    \expandafter\ifx\csname commentarrow\endcsname\relax$\leftarrow$\else \commentarrow[#2]\fi~%
    {\color{#2}##1}}%
  \expandafter\newcommand\csname #1st\endcsname[2]{%
    \colorbox{#2}{\color{white}\textsf{\textbf{#1}}}\sout{##2}~%
    \expandafter\ifx\csname commentarrow\endcsname\relax$\leftarrow$\else \commentarrow[#2]\fi~%
    {\color{#2}##1}}%
}
\newcommenter{TODO}{DodgerBlue1}
\newcommenter{rtron}{Green3}

\usepackage{comment}

\usepackage{pdfcomment}

\usepackage{soul}

\usepackage[normalem]{ulem}

\usepackage{csquotes}

\usepackage{tikz}
\usetikzlibrary{calc}
\usetikzlibrary{matrix}
\usetikzlibrary{chains}
\usetikzlibrary{shapes.geometric}
\usetikzlibrary{arrows.meta}
\usetikzlibrary{decorations.pathreplacing}
\usetikzlibrary{backgrounds}

\tikzset{
  dim above/.style={to path={\pgfextra{
        \pgfinterruptpath
        \draw[>=latex,|->|] let
        \p1=($(\tikztostart)!1.5em!90:(\tikztotarget)$),
        \p2=($(\tikztotarget)!1.5em!-90:(\tikztostart)$)
        in(\p1) -- (\p2) node[pos=.5,sloped,above]{#1};
        \endpgfinterruptpath
      }
    }
  },
  dim double above/.style={to path={\pgfextra{
        \pgfinterruptpath
        \draw[>=latex,|->|] let
        \p1=($(\tikztostart)!3em!90:(\tikztotarget)$),
        \p2=($(\tikztotarget)!3em!-90:(\tikztostart)$)
        in(\p1) -- (\p2) node[pos=.5,sloped,above]{#1};
        \endpgfinterruptpath
      }
    }
  },
  dim below/.style={to path={\pgfextra{
        \pgfinterruptpath
        \draw[>=latex,|->|] let 
        \p1=($(\tikztostart)!-1em!-90:(\tikztotarget)$),
        \p2=($(\tikztotarget)!-1em!90:(\tikztostart)$)
        in (\p1) -- (\p2) node[pos=.5,sloped,below]{#1};
        \endpgfinterruptpath
      }
    }
  },
}

\tikzset{
    right angle quadrant/.code={
        \pgfmathsetmacro\quadranta{{1,1,-1,-1}[#1-1]}     % Arrays for selecting quadrant
        \pgfmathsetmacro\quadrantb{{1,-1,-1,1}[#1-1]}},
    right angle quadrant=1, % Make sure it is set, even if not called explicitly
    right angle length/.code={\def\rightanglelength{#1}},   % Length of symbol
    right angle length=2ex, % Make sure it is set...
    right angle symbol/.style n args={3}{
        insert path={
            let \p0 = ($(#1)!(#3)!(#2)$) in     % Intersection
                let \p1 = ($(\p0)!\quadranta*\rightanglelength!(#3)$), % Point on base line
                \p2 = ($(\p0)!\quadrantb*\rightanglelength!(#2)$) in % Point on perpendicular line
                let \p3 = ($(\p1)+(\p2)-(\p0)$) in  % Corner point of symbol
            (\p1) -- (\p3) -- (\p2)
        }
    }
}

\newcommand{\pgfextractangle}[3]{%
    \pgfmathanglebetweenpoints{\pgfpointanchor{#2}{center}}
                              {\pgfpointanchor{#3}{center}}
    \global\let#1\pgfmathresult  
}

\usetikzlibrary{shapes.arrows}
\newcommand{\commentarrow}[1][Azure4]{\tikz[baseline=-3pt]{\node[shape border uses incircle, fill=#1,rotate=180,single arrow, inner sep=1pt, minimum size=6pt, single arrow head extend=2pt]{};}}

\tikzset{ax/.style={-latex,line width=2pt}}

\tikzset{camera/.style={fill=Sienna1,fill opacity=0.5},%
image plane/.style={draw=RoyalBlue3,line width=2pt}}

\usepackage{subfig}

\usepackage{graphicx}
\newcommenter{todo}{Firebrick1}
\newcommenter{dawei}{Blue}
\newcommenter{etank}{Purple}

\title{\LARGE \bf
  Stable Haptic Teleoperation of UAVs via Small $\cL_2$ Gain\\and Control Barrier Functions 
}

\author{Dawei Zhang$^{1}$, Roberto Tron$^{2}$
\thanks{$^{1}$Dawei Zhang is with the Department of Mechanical Engineering, Boston University, Boston, MA 02215, USA
        {\tt\small dwzhang@bu.edu}}%
\thanks{$^{2}$Roberto Tron is with the Department of Mechanical Engineering
 and the Division of Systems Engineering, Boston University, Boston, MA 02215, USA
        {\tt\small tron@bu.edu }}%
}

\begin{document}

\maketitle
% \thispagestyle{empty}
% \pagestyle{empty}

%%%%%%%%%%%%%%%%%%%%%%%%%%%%%%%%%%%%%%%%%%%%%%%%%%%%%%%%%%%%%%%%%%%%%%%%%%%%%%%%
\begin{abstract}

  We present a novel haptic teleoperation approach that considers not only the safety but also the stability of a teleoperation system. Specifically, we build upon previous work on \emph{haptic shared control}, which uses control barrier functions (CBFs) to generate a reference haptic feedback that informs the human operator on the internal state of the system, helping them to safely navigate the robot without taking away their control authority. Crucially, in this approach the force rendered to the user is not directly reflected in the motion of the robot (which is still directly controlled by the user); however, previous work in the area neglected to consider the feedback loop through the user, possibly resulting in unstable closed trajectories. In this paper we introduce a differential constraint on the rendered force that makes the system finite-gain $\cL_2$ stable; the constraint results in a Quadratically Constrained Quadratic Program (QCQP), for which we provide a closed-form solution. Our constraint is related to but less restrictive than the typical passivity constraint used in previous literature.  We conducted an experimental simulation in which a human operator flies a UAV near an obstacle to evaluate the proposed method.
\end{abstract}

%%%%%%%%%%%%%%%%%%%%%%%%%%%%%%%%%%%%%%%%%%%%%%%%%%%%%%%%%%%%%%%%%%%%%%%%%%%%%%%%
\section{INTRODUCTION}
Teleoperation allows human operators to remotely work in hard-to-reach or hazardous environments. 
When teleoperating an unmanned aerial vehicle (UAV), the limited field of view often leads to low levels of situational awareness, which can make it difficult to safely and accurately control the UAV \cite{mccarley2005human, Brandt2010}. To remedy these challenges, there mainly exist two orthogonal approaches. The first one is shared autonomy, where a supervisory controller modifies the inputs of the user to guarantee safety \cite{Brandt2010, Hou2013, zhang2020haptic}; these systems, however, reduce the control authority of the user. The second approach uses haptic signals to provide force feedback cues about the robot's behavior and the surrounding environment, which has been proven to help reduce dangerous collisions during teleoperation and improve operator situational awareness. However, these works mostly focus on improving safety, without considering the fact that the human operator will likely change the commanded input in response to the haptic cues, thus resulting in a closed feedback loop. Only few works considered the stability of the full human-robot-environment system \cite{stramigioli2010novel,rifai2011haptic,gioioso2015force}. In this paper, we propose a novel haptic teleoperation approach that considers not only the safety of the system but also the stability when designing the force-based haptic feedback. 
% In general, haptic teleoperation is composed of human operator, haptic device, controller, robot
% and environment
% as shown in Fig. \ref{control architecture}.

\subsection{Related work}\label{sec:related work}
In this section, we review previous work that designs force-based haptic feedback to help human operators navigate a robot. We briefly mention their main characteristics, Contrasting the novelty of our work in the next section.

Many researchers investigated algorithms about haptic feedback design. Haptic feedback that warns risk of collision is particularly relevant to the teleoperation of UAVs \cite{Lam2009, Brandt2010, zhang2020haptic}.  Lam et al. proposed a parametric risk field (PRF) to calculate the risk of a collision, which is the state-of-the-art approach \cite{Lam2009}. Brant and Colton set the magnitude of the force of the haptic feedback to be proportional to the time that it would take the UAV to collide with obstacles \cite{Brandt2010}. Recently, Zhang et al. designed an approach that uses control barrier functions (CBF) to generate haptic feedback that is based on the disagreement between the human's control input and the safe control input calculated by Control Barrier Functions \cite{zhang2020haptic}. However, these works mostly focus on the algorithmic design of the haptic feedback and lack a stability analysis of the teleoperation system. 

In this direction, there are few works designing architecture of the haptic teleoperation system and proving the stability of the system. Rifa{\"\i} et al. \cite{rifai2011haptic} used Lyapunov analysis to prove the input-to-state stability of the teleoperation loop. Similar to \cite{rifai2011haptic}, Omari et al. proved that the master system is input-to-state stable in the presence of bounded operator force and environment force \cite{Omari2013}. Most of the stability analysis has the assumption that the human operator will navigate the robot passively, and that the environment is dissipative \cite{mersha2013bilateral}.

Since passivity provides a sufficient condition for stability, making the system passive is an intuitive method to maintain the stability of a teleoperation system \cite{Niemeyer2008}. Lee \etal{} proposed a Passive-Set-Position-Modulation (PSPM) method that modulates the set-position signal to enforce the passivity of the system and applied PSPM to the haptic teleoperation of multiple UAVs to make the system passive over the Internet with varying-delay, packet-loss  \cite{lee2010passive, lee2011haptic}.

\subsection{Proposed system and contributions}
\begin{figure}[t]
\centering
\includegraphics[width=0.9\columnwidth,trim={0cm 0cm 0 0cm},clip]{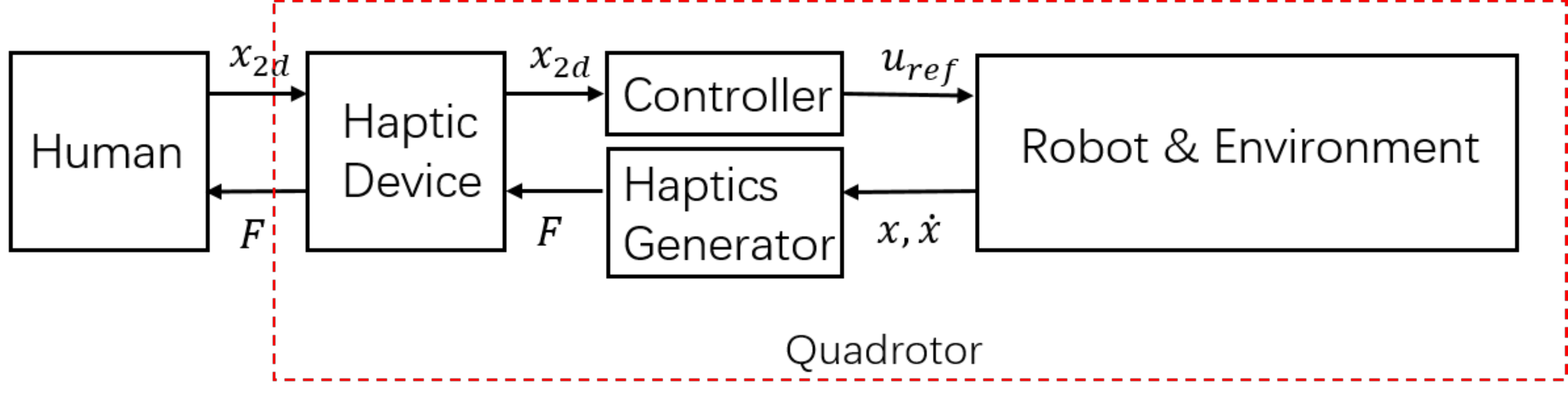}
\caption{\small Architecture of a haptic teleoperation system.}
\label{fig:control architecture}
\end{figure}

In this paper, we consider a teleoperation architecture of the form \ref{fig:control architecture}. The \emph{human} operator provides a desired velocity signal $x_{2d}$ for a robot (quadrotor, in our case) through a \emph{haptic device}. This desired velocity signal is given to a simple proportional velocity \emph{controller} that generates a reference control signal $u_{ref}$ which in turn is given to the actual \emph{robot}. The \emph{haptics generator} uses the state (position and velocity) of the robot to first compute a reference force $F_{ref}$ via a Control Barrier Function method, then passing a safe projected version $F$ which is rendered to the user via the haptic device. Note that the human and quadrotor subsystems form a closed-loop interconnection.

The key contribution of this paper lies in the design of a differential constraint for enforcing a finite $\cL_2$ gain from the user's input to the rendered force. This formulation leads to the following advantages:
\begin{itemize}
\item The $\cL_2$-grain differential constraint leads to a Quadratically Constrained Quadratic Program (QCQP), for which we provide a simple closed-form solution. 
\item Our method can be interpreted as a dynamic thresholding scheme that projects a desired reference force feedback to levels that are deemed to be safe (in the sense that they respect a desired $\cL_2$ gain).
\item Our approach does not assume that the force from the environment is passive, and can be applied to any scheme for generating the reference force. In this paper, we use the Control Barrier Functions method from \cite{zhang2020haptic}.
\item The new-designed differential constraint is less conservative than a similar constraint derived via strict output passivity; this translates to a better tracking of the desired reference haptic signal.
\end{itemize}

\section{PRELIMINARIES}
In this section, we give our main problem statement, and then review several concepts from control theory that will be used in the main body of the paper.
\subsection{Stability of teleoperation as a feedback interconnection}
We view the teleoperation architecture of Fig.~\ref{fig:control architecture} as a feedback connection of two subsystems, \emph{Human} and \emph{Quadrotor}, as shown in Fig.~\ref{fig:feedback_connection}. This interconnection is subject to two exogenous inputs: \emph{Human intention} represents the intentions of the operator (the desired motion), while \emph{Disturbance} represents unmodeled physical disturbances, such as gusts of wind or minor collisions. 

The goals of this paper are to design a force feedback scheme that ensures stability but that is also meaningful for the user, as formalized by the following:
\begin{goal}\label{prob:main}
  Design a \emph{haptic generator} map that guarantees bounded state trajectories of the system under bounded \emph{Human intention} and \emph{Disturbance} inputs and under suitable assumptions on the \emph{human} subsystem, 
\end{goal}
\begin{goal}\label{prob:characteristics}
 Design a \emph{haptic generator} which produces a force feedback $F$ with the following characteristics:
\begin{lenumerate}{C}
    \item \label{C1} If the quadrotor is far away from obstacles, or if the quadrotor is stationary, then $F=0$.
    \item \label{C2} The force is approximately proportional to the distance and the velocity of the quadrotor in the direction of the obstacle (the faster and the closer the quadrotor, the higher the expected force). If the robot is moving away from an obstacle, no force should be generated.
    \item \label{C3} The total amount of force received by the user should be bounded and approximately proportional for bounded inputs (i.e., if the user gives "small" commands, then also the force should be "small").
    \item \label{C4} Related to the previous point, the bounds on the output force should be applied over the entire trajectory, not at every time instant independently (in other words, the haptic generator should implement a map with some form of memory).    %The force can have some lag with respect to the input: if the quadrotor has non-zero velocity in the direction of the obstacle, there could be some force even if the input $x_{2d}$ is currently zero (i.e., there is not a static but a dynamic map between input $x_{2d}$ and output $F$).
\end{lenumerate}
\end{goal}

\subsection{Control Barrier Functions (CBFs)}\label{sec:review CBF}
Control Barrier Functions will be used to generate the reference haptic signal (force feedback).
\subsubsection{State Space Model}
Consider a dynamical system represented by the state space model
\begin{equation}\label{state model}
  \begin{aligned}
    \dot{x}&=f(x)+g(x)u\\
    y&=c(x)
\end{aligned}
\end{equation}
where $x\in\real{n}$ is the state of the system, $u\in\real{p}, y\in\real{}$ represent the vector of control inputs and the output, and $f: \real{n} \rightarrow \real{n}$, $g:\real{n}\rightarrow\real{n}\times\real{p}$, and $c:\real{n}\to\real{}$ are locally Lipschitz vector fields.
%$h: \real{n} \times \real{p} \rightarrow \real{p}$ is continuous, $f(0,0) = 0$, and $h(0,0) = 0$. The system has the same number of inputs and outputs.
\subsubsection{Lie derivatives}
We denote the Lie derivative of a function $h(x)$ along a field $f(x)$ as $L_{f} h(x)\defeq\frac{\partial h(x(t))}{\partial x(t)}\transpose f(x)$. %,  $L_{g} h(x) \defeq\frac{\partial h(x(t))}{\partial x(t)}\transpose g(x)$.
We denote with $L_{f}^{b} h(x)$ a Lie derivative of order $b$. The function $h$ has relative degree 2 with respect to the dynamics \eqref{state model} if $L_gh=0$, and $L_gL_fh$ is a non-singular matrix. In this case we have $\ddot{h}=L_{f}^{2} h(x)+L_{g} L_{f} h(x) u$.
\subsubsection{Safety Set}
A continuously differentiable function $h(x)$ can be define a  safety set $\cH$, as follows:
\begin{equation}\label{eqn:safety_set}
\cH: = \left\{ x \in\real{n} : h (x)\geq 0 \right\}. 
\end{equation} 
\subsubsection{CBFs for Second Order Systems}
The goal of control barrier functions is to produce a control field $u$ that makes a \emph{safe set} $\cH\subset\real{n}$ forward invariant, i.e., so that if $x(0) \in \cH $ then $x(t) \in \cH, \forall t>0 $ \cite{Ames2019}. Let $h(x)$ be a twice differentiable function representing $\cH$, i.e. $h(x)>0$ on the interior of $\cH$, $h(x)=0$ on its boundary, and $h(x)<0$ otherwise. Assuming that $h(x)$ has relative degree two, we can use a second-order exponential control barrier function \cite{Nguyen2016} to impose constraints on $u$ that ensure safety (i.e., forward invariance of $\cH$): 
\begin{multline}
 L_{f}^{2} h(x)+L_{g} L_{f} h(x) u 
+K\bmat{h(x)& L_{f}h(x)}^T \geq 0, 
\end{multline}
where $K\in\real{1\times 2}$ is a set of coefficients representing a Hurwitz polynomial.

\subsection{$\cL_2$ gain and feedback interconnections}
In this section we review concepts that will be at the center of our solution to Goal~\ref{prob:main}.
n
A map $\cC:u(t)\to y(t)$ between two signals has $\cL_2$-gain $k\geq 0$ if there exists a constant $\beta\in\real{n}$ such that $\norm{y}_2\leq k \norm{u}_2+\beta$.  Note that the map $\cC$ could be static (i.e., a simple function) or, more commonly, realized through a dynamical system.

The importance of this concept is given by the \emph{small gain theorem} (reproduced below in a slightly less generalized form specialized to our setting):

\begin{theorem}[\cite{khalil2002nonlinear}, Theorem 5.6, page 218] Assume  that and both systems are finite-gain $\cL_2$ stable with $\cL_2$ gains of $k_1$ and $k_2$: $\norm{u}_2 \leq k_1\norm{e_1}_2 + \beta_1$ and $\norm{F}_2 \leq k_2\norm{e_2}_2 + \beta_2$. If $k_1k_2<1$, then the feedback connection is finite-gain $\cL_2$ stable from the inputs $(\textit{Human Intention}, \textit{Disturbance})$ to the outputs $(e_1, e_2)$.
  
\begin{figure}[h]
\centering
\vspace{-10pt}
\includegraphics[width=0.8\columnwidth,trim={0cm 0cm 0 0cm},clip]{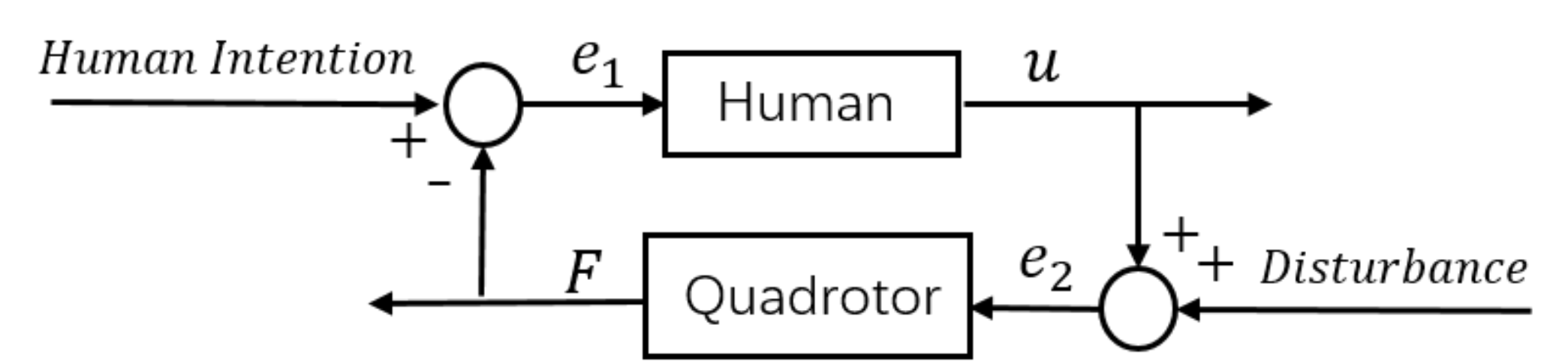}

\caption{\small Feedback connection.}
\vspace{-10pt}
\label{fig:feedback_connection}
\end{figure}
\end{theorem}

As it is common in the literature, we will model the human's reactions to the force feedback as a map with a finite $\cL_2$ gain.

\subsection{Passivity}
Although our final stability result will be based on the small gain theorem, passivity has been used to provide similar guarantees in previous work \cite{rifai2011haptic,lee2010passive}. We review the concept here for completeness; in Section~\ref{sec:methods} below we show that although one could use passivity to derive stability conditions similar to ours, these are significantly more restrictive. 
\begin{definition}
  The system \ref{state model} is said to be \emph{strictly output passive} if there exist a continuously differentiable positive semidefinite function $V(x)$ (called the storage function) and a static function $\rho(y)$ such that
\begin{equation}\label{strict out passivity constraint}
u\transpose y \geq \dot{V}+k y\transpose \rho(y)
\end{equation}
for all $(x,u)\in \real{n} \times \real{p}$, and $y\transpose\rho(y)>0$ for all $y\neq 0$.
\end{definition}
Intuitively, passivity states that an increase (or decrease) in the energy (storage function) of the system is upper bounded by the work ($u\transpose y$) that is possible to instantaneously transfer to (extract from) the system. 
A typical choice for the function $\rho$ is $\rho(y)=ky\transpose y$, which allows to connect passivity to $\cL_2$-gain theory:
\begin{lemma}[\cite{khalil2002nonlinear}, Lemma 6.5, page 242]
If the system \ref{state model} is output strictly passive with $u\transpose y \geq \dot{V} + ky\transpose y$, for some $k>0$, then its $\cL_2$ gain is less than or equal to $k\inverse$.
\end{lemma}
In our setting, this means that the mechanical energy that the user receives from the system will be limited by the energy of the input they provide divided by $k$.
% \begin{lemma}
% If the system \ref{state model} is passive with a positive definite storage function $V(x)$, then the origin of $\dot{x}=f(x,0)$ is stable.
% \end{lemma}
% \begin{definition}
% The system \ref{state model} is said to be zero-state observable if no solution of $\dot{x}= f(x,0)$ can stay identically in $S=\left\{x \in \real{n} \mid h(x, 0)=0\right\}$, other than the trivial solution $x(t) \equiv 0$. 
% \begin{lemma}
% Consider the system \ref{state model}. The origin of $\dot{x}= f(x,0)$ is asymptotically stable if the system is
% \begin{itemize}
%     \item strictly passive or
%     \item output strictly passive and zero-state observable
% \end{itemize}
% \end{lemma}
% \end{definition}
% \begin{theorem}
% The feedback connection of two passive systems is passive \cite{khalil2002nonlinear}.
% \end{theorem}
\subsection{Quadrotor dynamic model}
We consider a quadrotor that flies at relatively low speeds without highly aggressive maneuvers (which are exceedingly uncommon in a teleoperation setting), so that the roll and pitch angles of the quadrotor will remain small. Under such conditions, the dynamics of the UAV can be modeled by a double integrator, where the control input $u$ corresponds to the acceleration command of the UAV. Let $x=\smallbmat{x_1\\x_2}$ be the state of the quadrotor, where $x_1$ represents its position and $x_2 = \dot{x}_1$ its velocity. The dynamics of the system can be written as:
\begin{equation}\label{eq:double integrator}
\begin{array}{l}
    \bmat{\dot{x}_1\\\dot{x}_2} = \bmat{0&1\\0&0}\bmat{x_1\\x_2} + \bmat{0\\1}u,
\end{array}
\end{equation}
or, equivalently in matrix form:
\begin{equation}
    \dot{x}=Ax+Bu.
\end{equation}
\section{METHODS}\label{sec:methods}
To achieve Goal~\ref{prob:characteristics} and Goal~\ref{prob:characteristics}, we propose to design a three-steps haptic generator:
\begin{enumerate}
    \item Design a reference control input $u_{ref}$ that is based on the human user's control input $u$.
    \item  Generate a reference force $F_{ref}$ that guides the human user towards an input command that would be applied by a CBF-based collision-free controller.
    \item\label{it:stable force} Compute a force $F$ that is as close as possible to $F_{ref}$, but satisfies the characteristics listed in Goal~\ref{prob:characteristics} and Goal~\ref{prob:characteristics}.
\end{enumerate}

The rest of this section illustrates the details of each step of the force feedback design. For Step~\ref{it:stable force} we first discuss an alternative differential constraint based on passivity (Section~\ref{sec:passivity}), before basing to our proposed solution based on finite $\cL_2$ gain (Section~\ref{sec:finite gain}).
\subsection{Reference controller}
We define a simple reference proportional controller as
\begin{equation}\label{eq:uref}
    u_{ref}= \frac{1}{\Delta_t}(x_{2d}-B\transpose x)=\frac{1}{\Delta_t}(x_{2d}-x_2),
\end{equation}
where $x_{2d}$ is the input velocity set by the user, $B\transpose x=x_2$ is the current velocity of the robot,  and $\Delta_t$ is a time constant representing for how long $u_{ref}$ will be applied to the robot (i.e., $x_2$ will become $x_{2d}$ after $\Delta_t$, i.e., in a single step).

The dynamics of the quadrotor subsystem then becomes:
\begin{equation}
    \dot{x}=Ax+Bu_{ref}=(A-\frac{1}{\Delta_t}BB\transpose)x+\frac{1}{\Delta_t}B x_{2d}.
\end{equation}

We can rewrite the dynamics as:
\begin{equation}\label{new dynamics}
    \dot{x}=A_{new}x+B_{new}x_{2d},
\end{equation}
where $A_{new} = \bmat{0&1\\0&-\frac{1}{\Delta_t}}$ and $B_{new}=\bmat{0\\\frac{1}{\Delta_t}}$.

% \rtron{Note that this implies
% \begin{equation}
%     \dot{x}_2=\frac{1}{\Delta t} (x_{2d}-x_{2}).
% \end{equation}
% }
\subsection{Reference force}

We design the reference force in two steps as done in \cite{zhang2020haptic}. First, we compute the safe input $u_{CBF}$ that a CBF controller would provide for obstacle avoidance; then we design a reference force $F_{ref}$ that depends on the discrepancy between $u_{ref}$ and $u_{CBF}$. %; 
For the safe control input $u_{CBF}$ we apply the material reviewed in Section~\ref{sec:review CBF}:

\begin{equation}\label{eq:CBF-QP}
    \begin{array} { ll }
    {u_{CBF} =\underset{u \in \mathbb{R}^{m}}{\argmin}} {\frac{1}{2}\norm{u -u_{ref}}^{2}}\\
    \text { s.t. }L_{f}^{2} h(x)+L_{g} L_{f} h(x) u+K\bmat{h(x)& L_{f}h(x)}^T\geq 0.
    \end{array}
\end{equation}
where $f=Ax$ and $g=B$ are given by the original double integrator dynamics \eqref{eq:double integrator}.
% where $u_{ref}=\frac{1}{\Delta_t}(x_{2d}-x_2)$.

Then, we define the reference force $F_{ref}$ as:
\begin{equation}
    F_{ref}= u_{CBF}-u_{ref}.
\end{equation}

\subsection{Rendered force via passivity}\label{sec:passivity}
In this section we derive a differential constraint for designing the force $F$ based on strict output passivity. As shown in the experiments (Section~\ref{sec:experiment}) this approach gives inferior results, but it has been used in previous literature and represents a convenient stepping stone for explaining our approach.

\subsubsection{Energy design}

We first identify the storage function
\begin{equation}\label{eq:storage function}
    V(x)=\frac{k_v}{2}\norm{Bx}^2=\frac{k_v}{2}(x_2)^2,
\end{equation}
where $k_v$ is a constant parameter that adjusts the scale of the stored energy.

\subsubsection{Differential constraints}

We can find $F$ by looking for the force that is closest to $F_{ref}$ while satisfying the output passivity constraint:

\begin{equation}\label{eq:Passivity}
\begin{array} { c l } { \underset{F \in \mathbb{R}^{m}}{\argmin}}&{\frac{1}{2}\norm{F-F_{ref}}^{2}}\\
\text { s.t. } & x_{2d}\transpose F \geq \dot{V} + kF\transpose F,
\end{array}
\end{equation}
where we used the substitutions $u=x_{2d}$, $y=F$ in the strict output passivity constraint \eqref{strict out passivity constraint}.

\subsubsection{Stability}

Following Lemma 6.5 of Khalil, the derivative of $V$ satisfies
\begin{multline}\label{eq:passivity-L2 gain derivation}
    \dot{V}\leq u\transpose F-kF\transpose F=\\
    -\frac{1}{2k}(u-kF)\transpose (u-kF)+\frac{1}{2k}u\transpose u-\frac{k}{2} F\transpose F\\
    \leq \frac{1}{2k} \norm{u}^2-\frac{k}{2}\norm{F}^2.
\end{multline}
which implies
\begin{equation}\label{p_constraint}
    \frac{k}{2}\norm{F}^2\leq \frac{1}{2k} \norm{u}^2 - \dot{V}
\end{equation}
Integrating both sides we have
\begin{multline}
    \int_0^\tau \norm{F}^2 \de t \leq \frac{1}{k^2} \int_0^t\norm{u}^2\de t-\frac{2}{k} \int_0^t \dot{V}\de t\\=\frac{1}{k^2}\int_0^t\norm{u}^2\de t+\frac{2}{k} \bigl(V(0)-V(\tau)\bigr)\\
    \leq \frac{1}{k^2}\int_0^t\norm{u}^2\de t+\frac{2}{k} V(0)
\end{multline}
This shows that the quadrotor subsystem has $\cL_2$ gain equal to $k^{-2}$.
\subsubsection{Computational considerations}
% The RHS may less than 0 if we don't pick a proper parameter of $K_v$ in the storage function \eqref{eq:storage function}.
Problem \eqref{eq:Passivity} is a convex Quadratically Constrained Quadratic program, which, however, has a simple close form solution. To derive such solution, we use the quadrotor dynamics \eqref{new dynamics} to expand $\dot{V}=k_vx_2(x_{2d}-x_2)$, and then we rewrite the constraint \eqref{eq:Passivity} by completing the square: 
 % \begin{equation}
%     k\norm{F}^2-x_{2d}\transpose F \leq -\dot{V}.
% \end{equation}
%   or, equivalently,
 \begin{equation}
      \norm{F-\frac{x_{2d}}{2k}}^2 \leq \norm{\frac{x_{2d}}{2k}}^2 - \frac{k_v}{k}x_2\transpose \frac{1}{\Delta_t}(x_{2d}-x_2),
 \end{equation}
 \begin{equation}\label{eq:old_passivity}
   \norm{F-\frac{x_{2d}}{2k}}^2 \leq \frac{1}{4k^2}(x_{2d}\transpose x_{2d} -\frac{4kk_v}{\Delta_t}x_2\transpose x_{2d}+\frac{4kk_v}{\Delta_t}x_2\transpose x_2).
 \end{equation}
Requiring that the discriminant of the quadratic polynomial in the RHS of \eqref{eq:old_passivity} to be negative, we obtain that the constraint has a non-empty feasible region (i.e., positive RHS) under the condition that $0\leq\frac{kk_v}{\Delta_t}\leq 1$.

  With the constraint written in this form, we see that the QCQP problem \eqref{eq:Passivity} corresponds to a projection of $F_{ref}$ on the sphere centered at $\frac{x_{2d}}{2k}$ with radius given by the RHS of \eqref{eq:old_passivity}, which can be solved with simple geometrical considerations.
 
 Note that this closed-form solution highlights the main drawback of this passivity-based constraint: if the radius of the sphere is small, $F$ will be tied to be close to $\frac{x_{2d}}{2k}$, independently from $F_{ref}$; in other words, we might have a non-zero force even if $F_{ref}=0$, which is not desirable.

\subsection{Rendered force via finite gain}\label{sec:finite gain}
In this section we define a novel differential constraint that ensures a finite $\cL_2$ gain for the quadrotor subsystem. The intuition behind our main contribution is that strict output passivity is a sufficient but not necessary condition for a finite $\cL_2$ gain. This can be seen, for instance, from the fact that the inequality in \eqref{eq:passivity-L2 gain derivation} is, in general, not tight; instead, we directly start from \eqref{p_constraint}, but we also introduce an energy tank to balance the two sides of the equation, as described next.
\subsubsection{Energy design} For our approach, we use the same storage function $V(x)$ from \eqref{eq:storage function} that we used in the previous section. However, in order to make the constraint less restrictive, we also introduce an energy tank $E$ that is used to store energy when the reference force naturally satisfies \eqref{p_constraint}, and releases energy when the reference force violates that same constraint. Formally, we view $E$ as another state in the system, with dynamics
\begin{equation}
  \label{eq:tank dynamics}
  \dot{E}=\varepsilon.
\end{equation}

Note that we could also add a tank to the passivity-based approach from the previous section; nonetheless, in subsection \ref{subsec:tank energy limit } we show that we can impose $E(0)\equiv 0$ (i.e., the tank cannot store or release energy), which makes our approach comparable to the passivity-based method of Section \ref{sec:passivity}, but still superior in terms of performance, as shown in the experiments in Section \ref{sec:experiment}.
\subsubsection{Differential constraints}
We formulate a new force synthesis problem:
\begin{subequations}\label{eq:optimal F with tank}
    \begin{align}
    \min_{F,\varepsilon}& \frac{1}{2}\norm{F-F_{ref}}^2\\
      \subjectto&\frac{k}{2}\norm{F}^2+\varepsilon= \frac{1}{2k} \norm{u}^2 - \dot{V},\label{eq:finite gain tank}\\
    &\varepsilon\geq - \frac{E}{\Delta_t}\label{eq:Edot-nonnegative}
    \end{align}
  \end{subequations}
  where \eqref{eq:finite gain tank} is obtained by using the tank to balance \eqref{p_constraint}, and where \eqref{eq:Edot-nonnegative} imposes the fact that the energy tank cannot be depleted too fast (namely, in less than one time step $\Delta_t$). Additionally, note that \eqref{eq:Edot-nonnegative} also implies the constraint
  \begin{equation}\label{eq:Edot-condition}
    \varepsilon\geq 0 \textrm{ if } E=0.
  \end{equation}

\subsubsection{Stability}
Assuming $E(0)=0$ and integrating both sides of the constraint \eqref{eq:finite gain tank} we have
\begin{multline}
    \label{eq:integral-constraint}
    \int_0^\tau \norm{F}^2 \de t + \int_0^\tau \dot{E} \de t = \int_0^\tau \norm{F}^2 \de t + E(\tau)\\
    \leq \frac{1}{k^2}\int_0^t\norm{u}^2\de t+\frac{2}{k} V(0)
\end{multline}
which can be also rewritten as
\begin{multline}
    \label{eq:integral-constraint}
    \int_0^\tau \norm{F}^2 \de t \leq \frac{1}{k^2}\int_0^t\norm{u}^2\de t+\frac{2}{k} V(0) - E(\tau)
\end{multline}
Condition \eqref{eq:Edot-condition} implies $\int_0^\tau E \de t=E(\tau)\geq 0$, which, together with \eqref{eq:integral-constraint}, implies
\begin{equation}
    \label{eq:integral-constraint}
    \int_0^\tau \norm{F}^2 \de t 
    \leq \frac{1}{k^2}\int_0^t\norm{u}^2\de t+\frac{2}{k} V(0),
\end{equation}
which guarantees that the quadrotor subsystem has finite $\cL_2$ gain.

\subsubsection{Computational considerations}
Again, problem \eqref{eq:optimal F with tank} is a convex QCQP.
To find a closed-form solution, we obtain $\varepsilon$ from the equality constraint in \eqref{eq:optimal F with tank}, and rewrite the optimization problem as

\begin{equation}\label{eq:new form}
    \begin{aligned}
    \min_{F,\varepsilon}& \frac{1}{2}\norm{F-F_{ref}}^2\\
    \subjectto&\norm{F}^2 \leq \frac{2}{k}( \frac{E}{\Delta_t} +\frac{1}{2k} \norm{u}^2 - \dot{V}).
    \end{aligned}
\end{equation}
We can equivalently write the constraint of \eqref{eq:new form} as
\begin{equation}\label{eq:L2 gain sphere}
  \norm{F}^2 \leq \frac{1}{k^2}( \frac{2kE}{\Delta_t} +x_2\transpose x_{2d} -\frac{2kk_v}{\Delta_t}x_{2d}\transpose x_{2d}+\frac{2kk_v}{\Delta_t}x_2\transpose x_2).
\end{equation}
Similarly to the previous section, knowing that $E\geq 0$ and requiring that the discriminant of the quadratic form in the RHS to be negative, we obtain that the constraint has a non-empty feasible region (i.e., positive RHS) under the condition that $0\leq\frac{kk_v}{\Delta_t}\leq 2$.

With the constraint written in this form, we see that the QCQP problem \eqref{eq:new form} corresponds to a projection of $F_{ref}$ on the sphere centered at the origin with radius given by the RHS of \eqref{eq:L2 gain sphere}, which can be implemented with a simple thresholding on the norm of $F_{ref}$.

\subsubsection{Tank energy limits and comparison with passivity}\label{subsec:tank energy limit }
%We could also allow an "overdraft" of energy from the tank up to a point $E(\tau)\geq E_{\min}<0$, \rtron{or set $E(0)=E_{\min}>0$}in which case we get another constant $+\abs{E_{\min}}$ in the bound.

In practice, if the energy in the tank becomes too large, the bound on the force $F$ could become practically meaningless. Hence, we impose a threshold $E_{\max}$ on the maximum energy of the tank, and modify \eqref{eq:tank dynamics} to 
\begin{equation}
\dot{E}=\begin{cases}
\epsilon & \textrm{ if } E<E_{\max},\\
0 & \textrm{ otherwise.}
\end{cases}
\end{equation}

If we set $E_{\max}=0$, we essentially disable the energy tank; in this case the approach becomes directly comparable with the passivity-based approach. In both cases we obtain QCQP which can be solved by projections on spheres. Comparing the RHSs of \eqref{eq:L2 gain sphere} and \eqref{eq:old_passivity}, the radii of the two spheres are the same (up to a factor of $2$ in the choice of the coefficients). The main difference is that in the passivity approach the sphere is centered around $x_{2d}$, while in the proposed approach it is centered around the origin. As shown in the next section, the latter leads to a much more natural behavior.

\section{Experimental Validation}\label{sec:experiment}
\begin{figure}[t]
  \centering
  \includegraphics[width=\columnwidth, trim={0cm 0cm 0cm 0cm},clip]{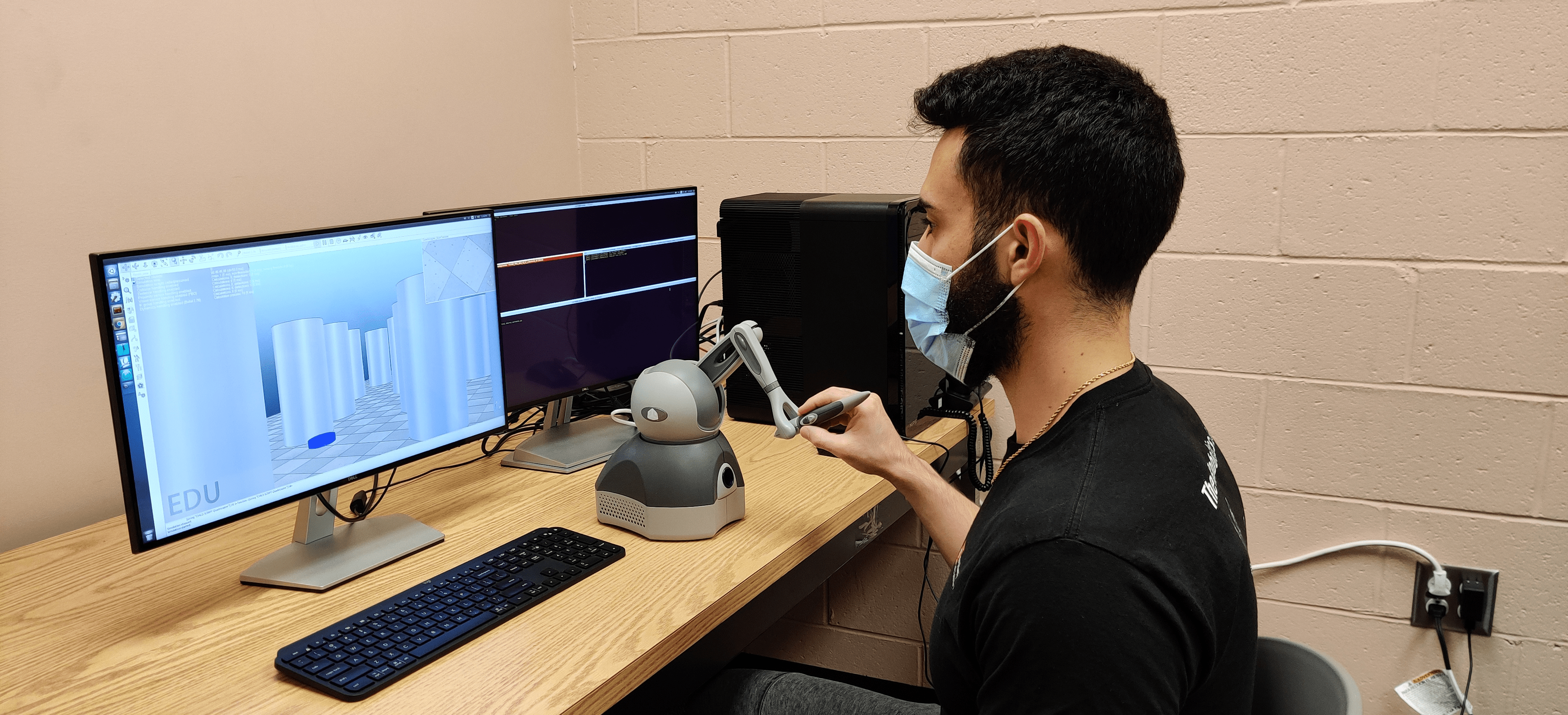}
  \caption{\small A human operator uses a haptic joystick to control the UAV in a simulated environment with a first-person view.}
  \label{fig:setup}
  \vspace{-10pt}
\end{figure}

\begin{figure}[t]
  \centering
  \includegraphics[width=\columnwidth]{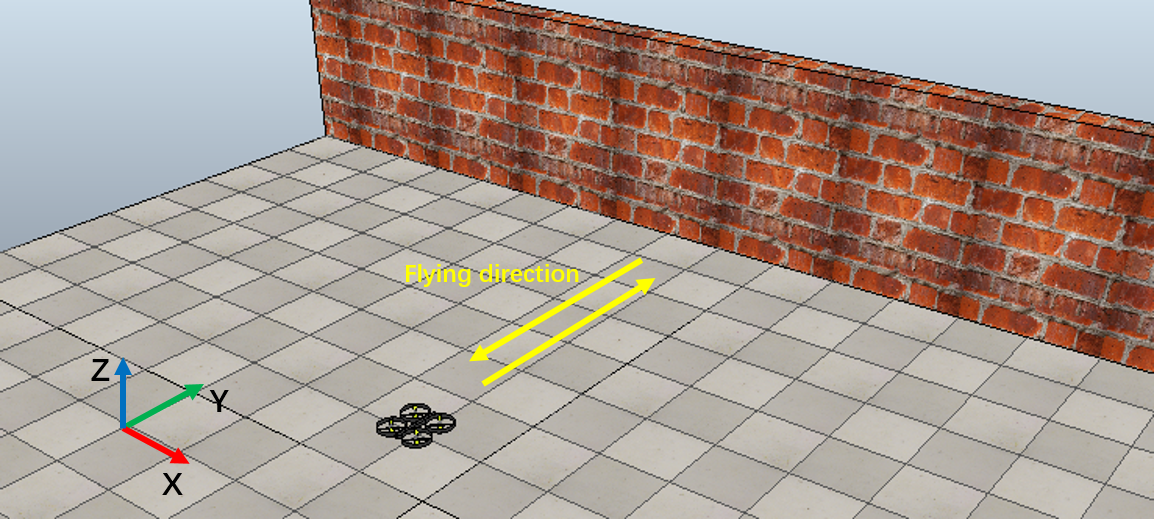}
  \caption{\small A quadrotor UAV is navigated to approach a wall.}
  \label{fig:overhead}
  \vspace{-10pt}
\end{figure}

\begin{figure}[h] 
  \centering
%   \vspace{-10pt}
  \subfloat[States of the UAV. \label{fig:result1}]{
    \includegraphics[width=0.9\columnwidth]{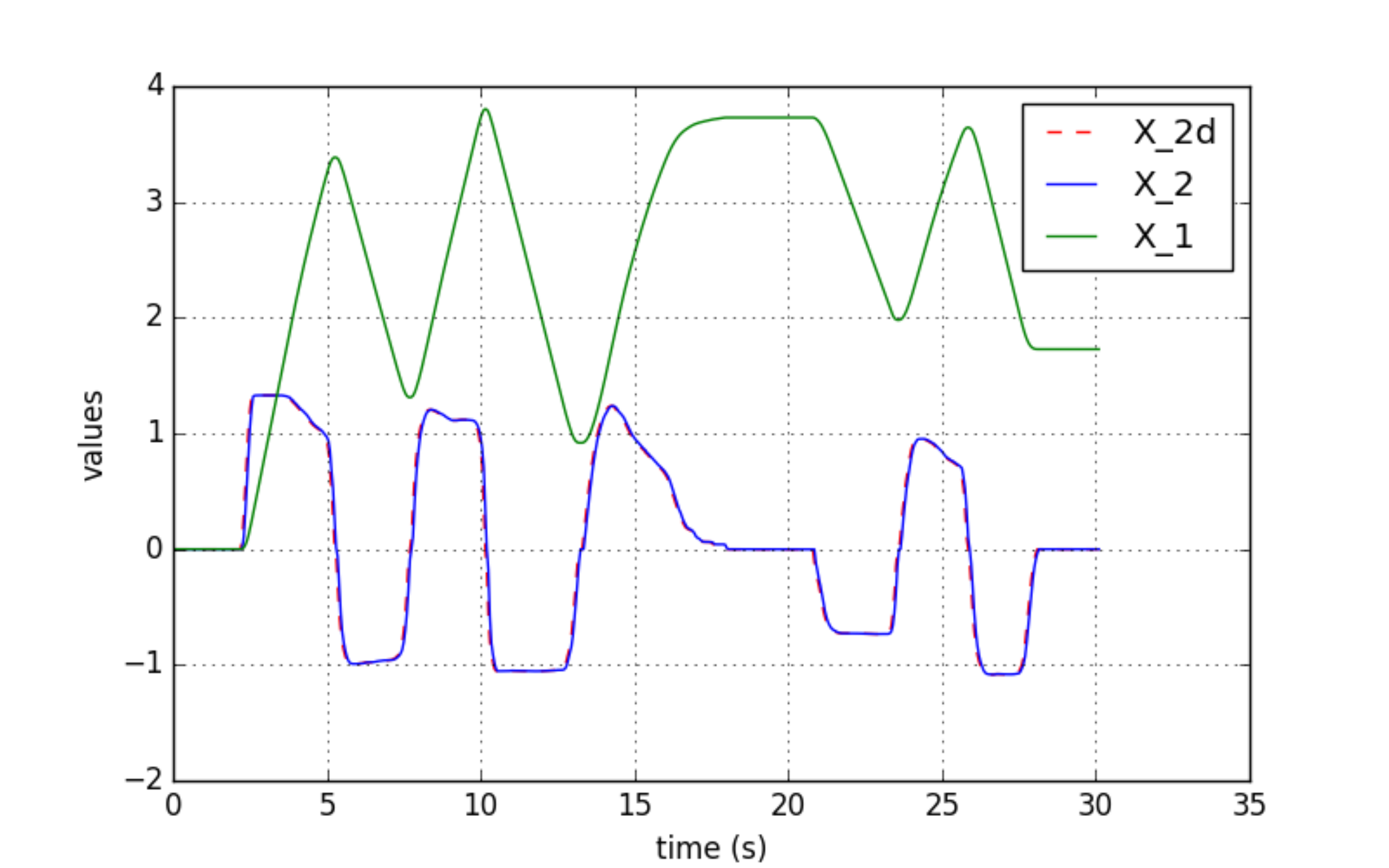}
    }
  \vspace{-10pt} 
  \subfloat[Force feedback with different methods. \label{fig:result2}]{
    \includegraphics[width=0.9\columnwidth]{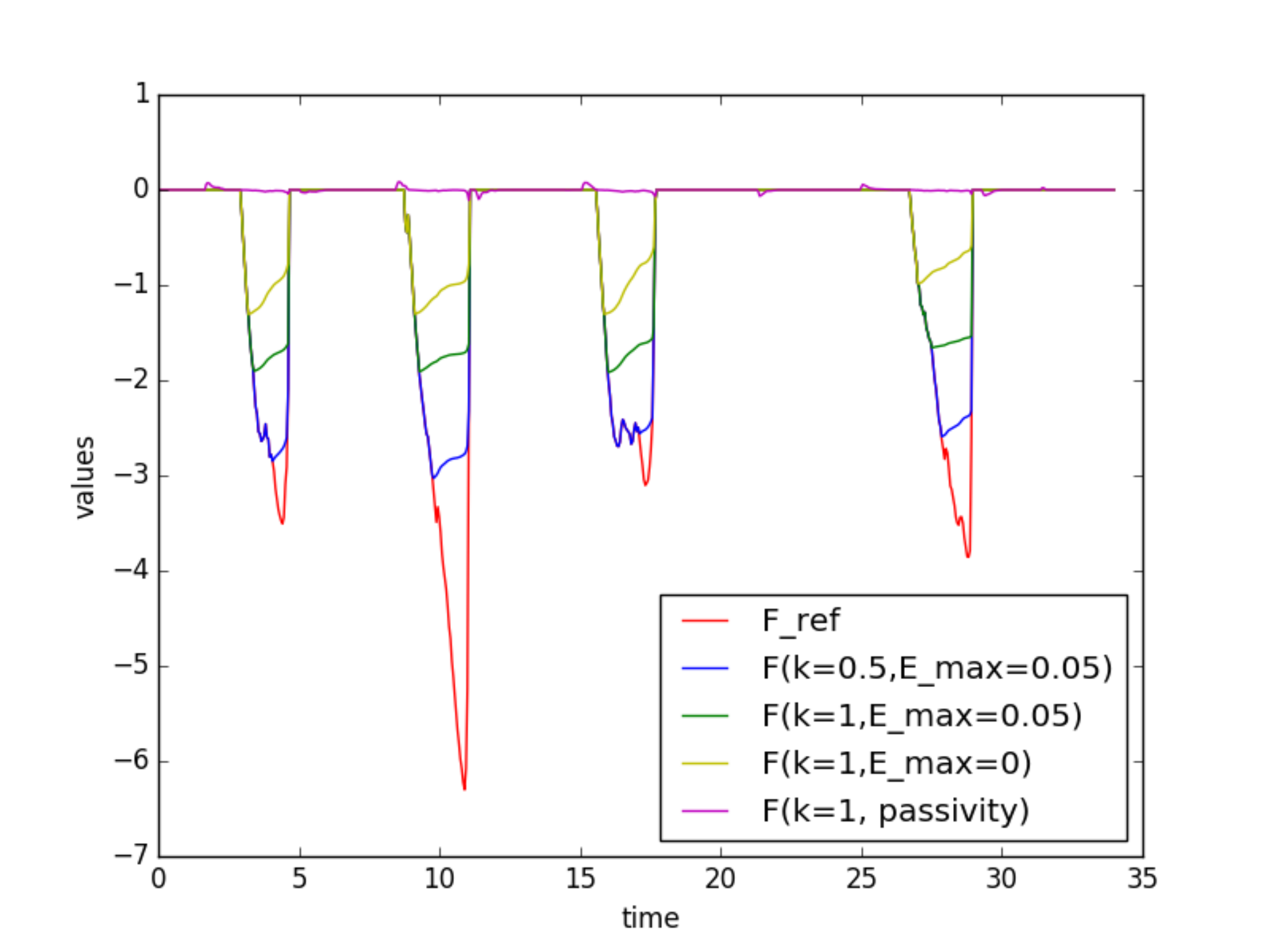}
   }
  \caption{\small Experimental results.}
  \label{fig:results}
\end{figure}

In this section, the proposed approach is evaluated through an experimental simulation in which the human operator navigates a simulated quadrotor in a virtual environment. 
\subsection{Experimental Setup}
The UAV and the environment are simulated using CoppeliaSim \cite{rohmer2013v}. As shown in Fig. \ref{fig:setup}, a 3D Systems Touch Haptic Device is used as the interface to control the motion of the UAV and provide haptic feedback to the operator. The communication between the haptic device and CoppeliaSim is performed via the Robot Operating System (ROS) middleware. The displacement of the stylus is mapped to the UAV's commanded velocity $x_{2d}$ through a constant of \unit[$0.2$]{$\frac{m/s}{cm}$}, with a dead-zone of \unit[$1$]{$cm$} to help the user give a control command with zero velocity. 

The experiment starts with navigating the UAV in a collision-free space. Then the human operator navigates the UAV towards and away from a vertical wall repeatedly for several times. 
During the experiment, we record the states of the UAV, the reference force feedback $F_{ref}$, and the projected force feedback $F$ that is perceived by the human operator in the y-direction shown in Fig. \ref{fig:overhead}.

As shown in Fig. \ref{fig:overhead}, the obstacle in this experiment is a vertical wall that is \unit[$4$]{$m$} away from the starting position of the UAV. Therefore, we pick the CBF in the form of
\begin{equation}
    h(x) = A_h\transpose x + B_h,
\end{equation}
where $A_h =\bmat{0\\-1} $, $B_h = 4$.
In this simulation, we set up $\Delta_t$ as \unit[$0.05$]{$s$} and $k_v$ in the storage function \eqref{eq:storage function} as $\frac{\Delta_t}{2k} = \frac{0.025}{k}$.

\subsection{Results and Discussion}
We plot the results of the experiments projected on the $y$-axis coordinate in Fig. \ref{fig:results}. As we can see from Fig. \ref{fig:result1} and Fig. \ref{fig:result2}, the force feedback that is provided to the human operator is zero when either the UAV flying away from the wall (e.g. from \unit[$11$]{$s$} to \unit[$13$]{$s$}) or the UAV staying stationary (e.g., \unit[$18$]{$s$} to \unit[$20$]{$s$}), which satisfies the proposed characteristic \ref{C1}. The CBF generates a reference force feedback $F_{ref}$ as the UAV approaches the wall fast or gets close to the wall (e.g., from \unit[$8$]{$s$} to \unit[$10$]{$s$}) while the force feedback has the same trend with $F_{ref}$, which indicates that \ref{C2} is satisfied. Furthermore, the value of the force feedback is bounded by the human's input with a $\cL_2$ gain $k$, as depicted in Fig. \ref{fig:result2}. A smaller value of $k$ leads to a greater value of the force feedback. When the discrepancy between the human's control input and the CBF's safe input is too large, the bounded force feedback will keep the system finite gain stable. This result is consistent with the characteristic \ref{C3}. As we can see from the result with condition ($k=1$, $E_{max} =0$) and condition ($k=1$, $E_{max} =0.05$), the bounds of the force feedback decrease over time (e.g., from \unit[$16$]{$s$} to \unit[$17$]{$s$} ), which aligns well with the expected \ref{C4}. In addition, when applying the energy tank, the human operator receives a relatively higher value of the force feedback which depends on the upper limit, $E_{max}$, of the energy tank. As shown in Fig. \ref{fig:result2}, when comparing the result under the condition ($k=1$, $E_{max} =0$) with the result under the condition ($k=1$, $E_{max} =0.05$), we can find that allowable force feedback can be increased by increasing $E_{max}$. Also, we can conclude that our approach with the energy tank is less conservative than the method without the energy tank and the method via strict output passivity.

\section{CONCLUSIONS AND FUTURE WORK}
In this paper, we proposed a novel haptic teleoperation approach that uses control barrier functions and small $\cL_2$ gain to maintain not only the safety but also the stability of the full human-robot-environment system. We conducted an experimental simulation in which a human operator flies a UAV near an obstacle to evaluate the proposed method. The results show that the proposed approach behaves very similarly to a simple thresholding of the force generated by the CBF-based haptic method, and satisfies all the characteristics that we would have expected by an intuitive haptic teleoperation interface. 

In this work, we investigated our approach under the \emph{haptic shared control} paradigm in which the human operator always keeps the control authority of the robot. In the future, we will further investigate our approach in a \emph{haptic shared autonomy} paradigm where the human's control command to the robot is modified by CBF.
\bibliographystyle{IEEEtran} 

\bibliography{biblio/IEEEfull,biblio/IEEEConfFull,biblio/OtherFull,% Do not insert spaces in this command, otherwise it will not work.
  biblio/tron,%
  biblio/formationControl,%
  biblio/websites,
  biblio/dawei}

\end{document}